\documentclass{article}

\usepackage[preprint]{neurips_2023}

\usepackage[utf8]{inputenc} 
\usepackage[T1]{fontenc}    
\usepackage{hyperref}       
\usepackage{url}            
\usepackage{booktabs}       
\usepackage{amsfonts}       
\usepackage{nicefrac}       
\usepackage{microtype}      
\usepackage{xcolor}         

\usepackage[pdftex]{graphicx}

\title{Reinforcement Learning from LLM Feedback to Counteract Goal Misgeneralization}

\author{
  Houda Nait El Barj\\
  Stanford University\\
  \texttt{hnait@stanford.edu} \\ 
  \And
  Théophile Sautory\\
  UC Berkeley \\
  \texttt{tsautory@berkeley.edu } \\
}

\begin{document}

\maketitle

\begin{abstract}
We introduce a method to address goal misgeneralization in reinforcement learning (RL), leveraging Large Language Model (LLM) feedback during training. Goal misgeneralization, a type of robustness failure in RL occurs when an agent retains its capabilities out-of-distribution yet pursues a proxy rather than the intended one. Our approach utilizes LLMs to analyze an RL agent's policies during training and identify potential failure scenarios. The RL agent is then deployed in these scenarios, and a reward model is learnt through the LLM preferences and feedback. This LLM-informed reward model is used to further train the RL agent on the original dataset. We apply our method to a maze navigation task, and show marked improvements in goal generalization, especially in cases where true and proxy goals are somewhat distinguishable and behavioral biases are pronounced. This study demonstrates how the LLM, despite its lack of task proficiency, can efficiently supervise RL agents, providing scalable oversight and valuable insights for enhancing goal-directed learning in RL through the use of LLMs.
\end{abstract}

\section{Introduction}
In recent years, there has been a growing concern about the possibility of a catastrophic future where humanity is disempowered by AI (\cite{chan_2023}, \cite{bostrom_superintelligence}, and \cite{amodei_2016}). Such a catastrophic scenario supposes that AI is misaligned, in the sense that it has the capacity to overtake humans while pursuing unintended  and undesirable goals which do not match those humans originally aimed for it (\cite{gabriel_2020}, \cite{hendrycks_2023}). Over the past decades, the capabilities of AI systems have been expanding in virtually every domain: playing board games like Go (\cite{silver_2016}), video games (\cite{vinyals_2019}), skin cancer diagnosis (\cite{esteva_2017}) or protein structure prediction (\cite{jumper_2021}). 
Thus it could soon surpass human performance on a wide range of tasks. Consequently, tackling the challenge of alignment seems more pressing than ever. 

In the field, researchers distinguish between two modes of alignment: 'outer alignment,' which involves creating a reward function that accurately reflects human preferences and goals, and 'inner alignment,' which focuses on ensuring that a policy trained on such a function adheres to these preferences. Outer misalignment occurs due to inaccurately specified rewards, whereas inner misalignment typically stems from the model's inductive biases. 
A large body of research documents outer misalignment, showing how AI systems can exploit misspecified reward functions (\cite{pan_2022}). 
This includes 'specification gaming,' where AI manipulates the reward generation process (\cite{krakovna_2020}), and 'reward hacking,' involving exploiting loopholes in task specifications (\cite{skalse_2022}, \cite{leike_2017}, \cite{everitt_2017}, \cite{ibarz_2018}).
Solutions to address reward misspecifications have been discussed in the literature, with reinforcement learning with human feedback (RLHF) gaining large traction (\cite{christiano_2017}, \cite{ziegler_2020}, \cite{schulman_2017}). This approach, which trains models using reward functions derived from human feedback, guides the agent's behavior and reduces dependence on potentially flawed hand-crafted reward functions, although it may still sometimes lead to reward hacking (\cite{amodei_2017}, \cite{stiennon_2022}).

On the contrary research on inner alignment, and goal misgeneralization has been slower. Goal misgeneralization occurs when a model acts to achieve an unintended goal, despite receiving accurate training feedback. It arises as the model is unable to generalize the intended goal to new environments, and is a type of inner alignment failure. The relative lack of research in this area can be explained by a few reasons. First, instances of goal misgeneralization are hard to detect. They require to test the policy on  distributions that varies significantly from training, thus yielding little practical examples. Furthermore, goal misgeneralization is likely to become more of an issue as AI complete more complex tasks which is not a direct concern today (unlike tackling reward misspecification). Finally, a rigorous framework of goal misgenerelization would require carefully defining which conceptual unintended goal the AI system is pursuing. This would require more research in interpretability technique, although it has been a growing and impactful field lately (\cite{wang_2022}, \cite{meng_2023}, \cite{anthropic_2023})In complex environments where goal misgeneralization is more likely, scalable oversight becomes a critical concern in research.Therefore, developing methods that minimize human involvement in training and evaluation is essential (\cite{bowman_2022}).

In this paper, we introduce a novel scalable method to tackle goal misgeneralization in reinforcement learning (RL) agents. Our approach integrates a Large Language Model (LLM) with RL training. The LLM assists by assessing the agent's policies and suggesting training modifications to mitigate the risk of goal misgeneralization. We also derive a reward model from the LLM preferences and feedback. Our method helps counteract biases learned by the RL agent. This method is particularly relevant in scenarios where the environment's reward function is not well-defined, or alterations to the training environment are not feasible.
We evaluate our method on a maze navigation task. We show that despite the LLM's inability to perform the task itself, it effectively supervises the RL agent in learning its goal robustly. Our results demonstrate a significant reduction in goal misgeneralization instances across a variety of settings. Finally, we discuss how various parameters influence our method's effectiveness.

Our main contributions are as follows:
\begin{itemize}
    \item We provide a framework to counteract goal misgenelarization in RL agents without explicitly training the agent on out-of-distribution data.
    \item We exhibit an RL application where a reward model is derived from LLM preferences, enabling LLMs to efficiently supervise RL agents despite not being proficient in the task.
\end{itemize}

\section{Preliminaries} 
In this section, we  review the definition of goal misgeneralization introduced in \cite{langosco_2023} and provide an overview of our method. 

\subsection{Goal Misgeneralization}
\cite{langosco_2023} define goal misgeneralization for a reinforcement learning agent as an instance where a learnt policy $\pi$ achieves a low test reward even though it was achieving a high training reward. The policy appears to be optimizing a different reward than the one originally trained for. More precisely, let $G$ be our intended goal. Consider the case where the agent is deployed in a new environment that presents a distributional shift. Then, $\pi$ is subject to goal misgeneralization if the reward appears to be optimizing a goal $G^\prime \neq G$ in the new test environment. We call  $G^\prime$  the behavioral objective of the agent. In this new  environment, the policy continues to act capably yet achieves a low reward.
For goal misgeneralization to happen, it must be case that either the training environment is not diverse enough or that that there exists some proxy that correlates with the intended objective during the training, but comes apart once the agent is deployed in the test environment.

\subsection{Reward modeling}
Consider a set $D$ of preferences over policy rollouts. Each element of $D$ is triple $(\omega^1, \omega^2, \mu)$ where $\omega^1$ and $\omega^2$ are the two policy roll-outs in a pair, and $\mu$ is a distribution over $\{1,2\}$ indicating which roll-out is preferred.  $\mu$ concentrates its mass entirely on the preferred choice or remains uniform if there's no preference or equal preference for both options. 
Following \cite{christiano_2017}, we train a reward model $\mathcal{R}$ from the set $D$ by minimizing the cross-entropy loss between the predicted probabilities $\hat{P}$ and the preference labels $\mu$:
$$\mathcal{L}(\mathcal{R}) = - \sum_{(\omega^1,\ \omega^2,\ \mu) \in D} \mu(1) \log \hat{P}(\omega^1 \succ \omega^2) + \mu(2) \log \hat{P}(\omega^2 \succ \omega^1) $$
where 
$$\hat{P}(\omega^1 \succ \omega^2) = \frac{\exp \sum_{\omega^1} \mathcal{R}(s_t, s_{t+1})}{\exp \sum_{\omega^1}\mathcal{R}(s_t, s_{t+1})+ \exp \sum_{\omega^2} \mathcal{R}(s_t, s_{t+1})}.$$

\subsection{RLAIF agent}
We consider an RL agent that is prone to goal misgeneralization. This agent is trained for $\tau \cdot T$ timesteps with a specific reward function $R$. While this reward function correctly captures the intended goal during the training, we expect the agent to fail to generalize its goal in out-of-distribution environments because of the existence of a proxy between the intended goal and the agent's learnt goal. We rely on a Large Language Model (LLM) to identify and correct these failure modes by analyzing the behavior of the RL agent. In particular, we train the agent using a reward model derived from the LLM feedback.  The goal of this fine-tuning period is to increase the agent's generalization capability. The full methodology of this reinforcement learning from LLM feedback is detailed in the next section.

\section{RL from LLM Feedback to reduce Goal Misgeneralization}
In this section, we describe the steps of the reinforcement learning with LLM supervision. 

\subsection{Initial RL training}
The RL agent is trained for $\tau\cdot T$ timesteps on an environment prone to goal misgeneralization.  At $\tau\cdot T$ timesteps we pause the training and sample $N^{I}$ policy roll-outs $\omega^{\pi}$. A roll-out $\omega^{\pi}$ is a sequence of state-action pairs generated by the agent's policy $\pi$ at time $\tau \cdot T$.

\subsection{LLM assessment and suggestions} \label{sec:method_llm_assessment}
Similarly to how parents observe their children behaviours in different environment and then implement different education methods to correct them, we rely on the assistance of an LLM parent to correct the biased behavior of our RL agent. The objective of LLM supervision is to enable the RL agent to not only accurately generalize its goals but also to maintain competent performance. We input the  $N^{I}$ samples into the LLM, prompting it to analyze potential scenarios where our current training setup might cause the agent to inadequately generalize its goal across diverse testing environments. The input to the LLM is structured as follows :

\begin{enumerate}
    \item \textit{Preamble} A description of the RL problem including the objective of goal generalization
    \item \textit{Request for analysis} A question to elicit the LLM into thinking about how the current setup could fail goal generalization 
     \item \textit{Explanation of the setup} Description of the environment 
    \item \textit{Policy rollouts} Policy rollout samples 
\end{enumerate}

The LLM then suggests modification to the training environment that would mitigate goal misgeneralization.  At this point, a straightforward solution to implement the LLM feedback would be to continue the training on the suggested diversified environments directly. Nonetheless, we hypothesize a context where the underlying reward function of these environments cannot be accessed or where the training environment cannot be adjusted. Hence, we aim to correct the agent's behavior after it has learnt a biased policy preventing it from generalizing its goal. In order to correct this bias, we plan to train our agent using a reward model induced from the LLM preferences. We generate the LLM's suggested environments and deploy our agent on these environments. We sample $N^{R}$ policy roll-outs. The sole purpose of this training is to construct the LLM-informed reward model.

\subsection{LLM preference labeling}
From the $N^R$ policy roll-outs sampled, we form a set comprising all possible pairs. We give the LLM a prompt and two candidate policy roll-outs and and instruct it to rate which policy roll-out is preferred. The input to the LLM for this preference labeling is structured as follows: 

\begin{enumerate}
    \item \textit{Preamble} A description of the RL problem including the objective of goal generalization
    \item \textit{Indication of reward model (optional)} An indication that the LLM's answer will be integrated to a reward model for the RL training
     \item \textit{Explanation of the maze textual representation} A key  for the textual representation of the maze  
    \item \textit{Policy rollouts} Policy rollout samples in their textual representation
    \item \textit{Ending} An ending string to ensure the LLM answer is in a fixed format (e.g. "Preferred traejctory=") 
\end{enumerate}

The LLM preferences are stored in a database $D$ of triples $(\omega^1, \omega^2, \mu)$ where $\omega^1$ and $\omega^2$ are the two policy roll-outs in a pair, and $\mu$ is a distribution over $\{1,2\}$ indicating which roll-out the LLM preferred. If the LLM has a preference, then $\mu$ puts all its mass on that choice. On the contrary, if none is preferred, or if both are equally preferable, then $\mu$ is uniform.

\subsection{Reinforcement learning from LLM preferences}
\label{sec:method_llm_reward_model}
We train a reward model $\mathcal{R}$ from the set $D$ of LLM-preferences. 
We then resume the training of the RL agent on the training environment. We experiment with combining the LLM-induced reward model with the original reward function. The composite reward function is implemented as:
$$ R^{'} = \lambda \cdot \mathcal{R} + (1-\lambda) \cdot R$$
In this case, implementing our method fully correspond to $\lambda =1$. 

\section{Dataset and experiment details}

\subsection{Dataset}
We use the OpenAI Procgen environment suite \citep{openaiprocgen}, which  has been designed to study RL agents generalization capacities for specific tasks.
We choose the Maze game environment where a mouse is trained to navigate through a maze towards a randomly located cheese. A reward function is integrated within the maze game environment and provides a reward of $10$ when the agent reaches the cheese's location, and a reward of $0$ for any other action.

\subsection{Goal Misgeneralization in Procgen Maze Game}
\begin{figure}[h!]
  \centering
  \includegraphics[width = 0.4 \textwidth]{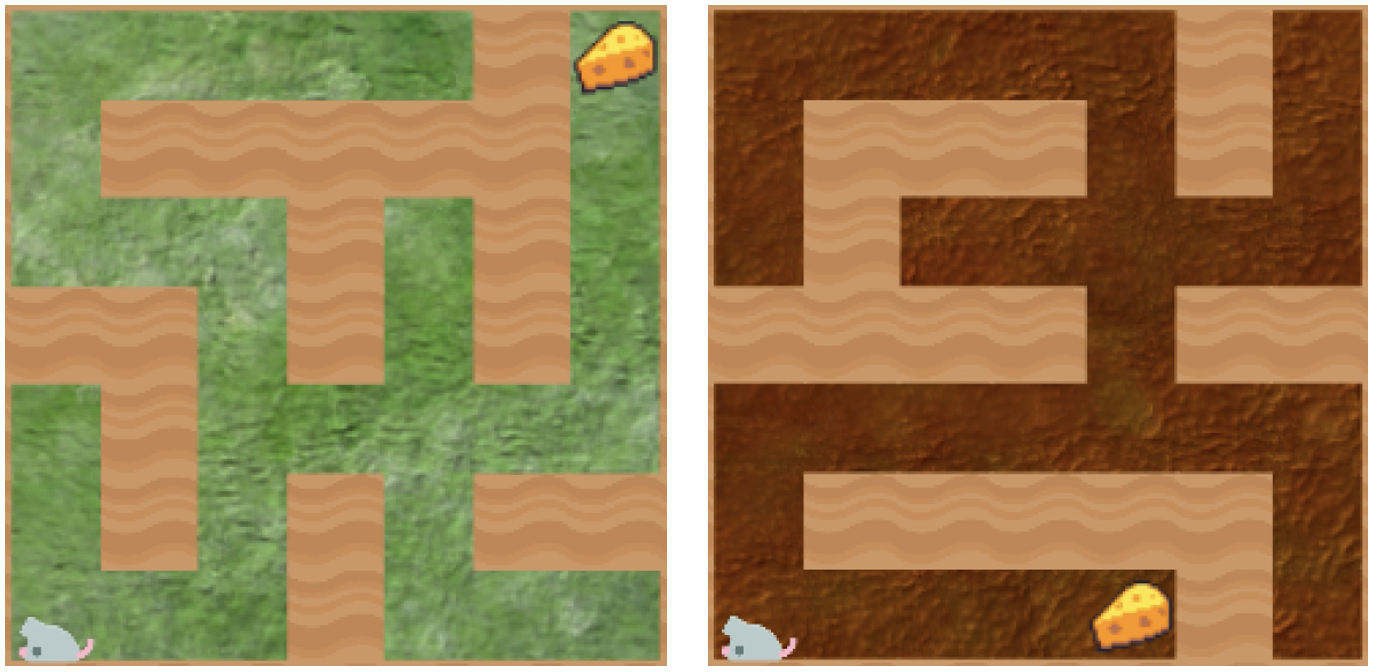}
  \caption{Example of cheese location variants. The left hand side shows an example of the fixed goal position during training. The right hand side is an example of the random test position.}
  \label{fig:mazes_cheese_location}
\end{figure}

We follow \cite{langosco_2023} setup to create a setting in the maze game that is prone to goal misgeneralization. In particular, we randomly locate the goal within a region of size $1$ to $10$ in the upper right corner of the maze during the training environment. However, in the test environment, the goal is randomly located in any region within the maze (see  figure \ref{fig:mazes_cheese_location}). \\

We follow a zero-shot protocol in all experiments: the agent does not see the out-of-distribution testing environment during training. \cite{langosco_2023} demonstrated that RL agents trained with the reward function in environments where the goal is consistently placed in the upper right corner learn to navigate there. The learnt objective $G'$ is the position (upper right) rather than the intended objective $G$, which is the cheese. However, as the randomization region grows, accurate goal generalization becomes more likely. 

\subsection{RLAIF training on the Procgen Maze game}
\begin{figure}[h!]
  \centering
  \includegraphics[width = 0.4 \textwidth]{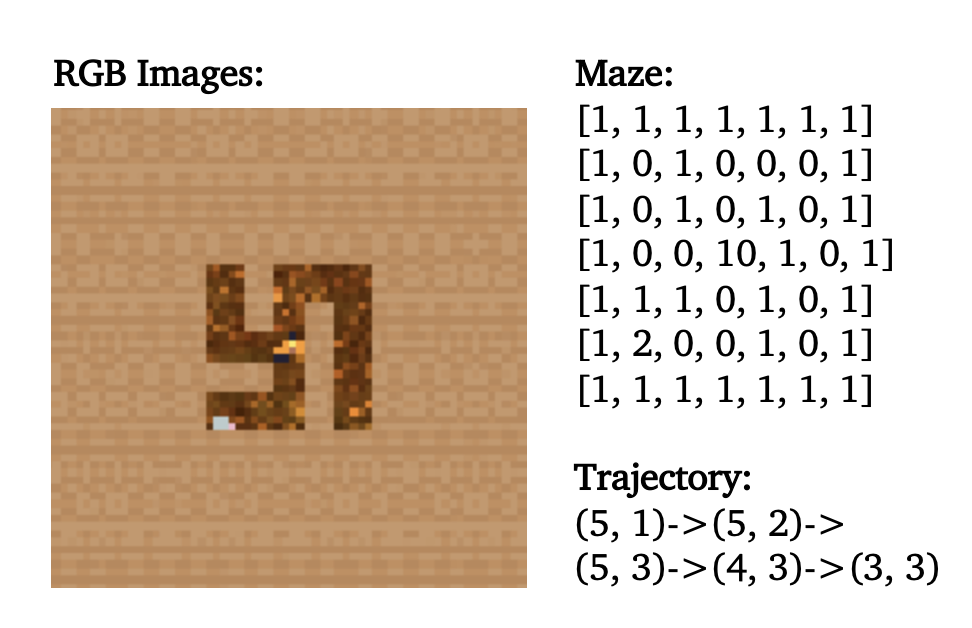}
  \caption{An example of a roll-out representation. The RGB image is the environment. The maze is its textual representation neglecting the wall padding. The trajectories are the steps taken by the agent, where (0, 0) is the upper left corner of the maze.}
  \label{fig:data_rollout}
\end{figure}

\paragraph{RL Agents} We train two RL agents in this setup for $T=75M$ timesteps in total. One agent is trained for the full period with the environment reward function. This replicates the results of \cite{langosco_2023}. This base agent acts as a benchmark for evaluating our method. The other agent is trained according to our methodology. This fine-tuned agent is trained for $\tau= \frac{2}{3}$ \% of the $75M$ timesteps with the reward function. We then pause its training and sample $N^{I} = 300$ policy roll-outs. 
\paragraph{LLM choice and suggestions}
The policy roll-outs are stored as a sequence of RGB images alongside the corresponding textual representation of the environment and trajectory (see Figure \ref{fig:data_rollout} for an example). 

\begin{figure}[h!]
  \centering
  \includegraphics[width = 1 \textwidth]{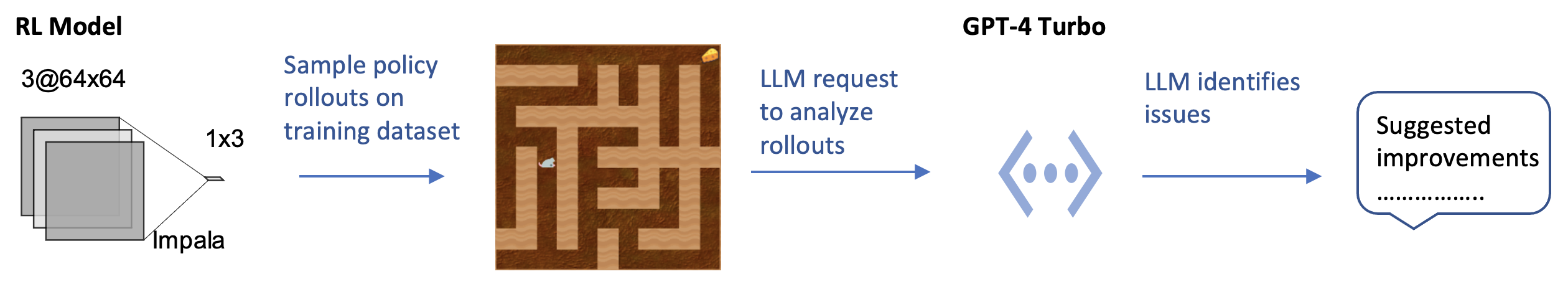}
  \caption{LLM assessment and suggestions. Policy rollouts of the agent are sampled on the training dataset. They are shared with the LLM which identifies situations in which the current policy could fail, and suggest improvements to the dataset.}
  \label{fig:methodology_step1}
\end{figure}

We use GPT-4 Turbo (GPT-4) as the parent-LLM.
We input the policy rollout samples into GPT-4, prompting it to analyze potential scenarios where our current training setup might cause the agent to inadequately generalize its goal across diverse testing environments. 
An example of the input to GPT-4 is provided in figure  \ref{fig:prompt_env_analysis}.  In 80$\%$ of our queries, GPT-4  is able to identify the confoundedness in the  environment by clearly indicating that the location and the cheese are undistinguishable (see an example in Figure \ref{fig:gpt_env_analysis}). Note that GPT-4 also identifies other potential flaws in the current set up that could lead to failure of generalization. For the scope of our paper, we do not include these further points. A summary of this initial step is provided in figure \ref{fig:methodology_step1}.

\paragraph{Preference labeling and reward model} Given GPT-4's suggestion of varying the location of the cheese in training, we create a new environment where the cheese is located randomly in all parts of the maze. We deploy our agent on this new environment, and sample $N^{R}=158$ policy roll-outs. The sole purpose of this evaluation is to construct the LLM-informed reward model.
From the $158$ policy roll-outs sampled, we form a set comprising all possible pairs with the condition that all mazes in a given pair have identical dimensions. This enhances the informativeness of the preference labeling by ensuring that variations in maze size do not influence the LLM's choices. We form a total of $1800$ pairs.

\begin{figure}[h!]
  \centering
  \includegraphics[width = 1 \textwidth]{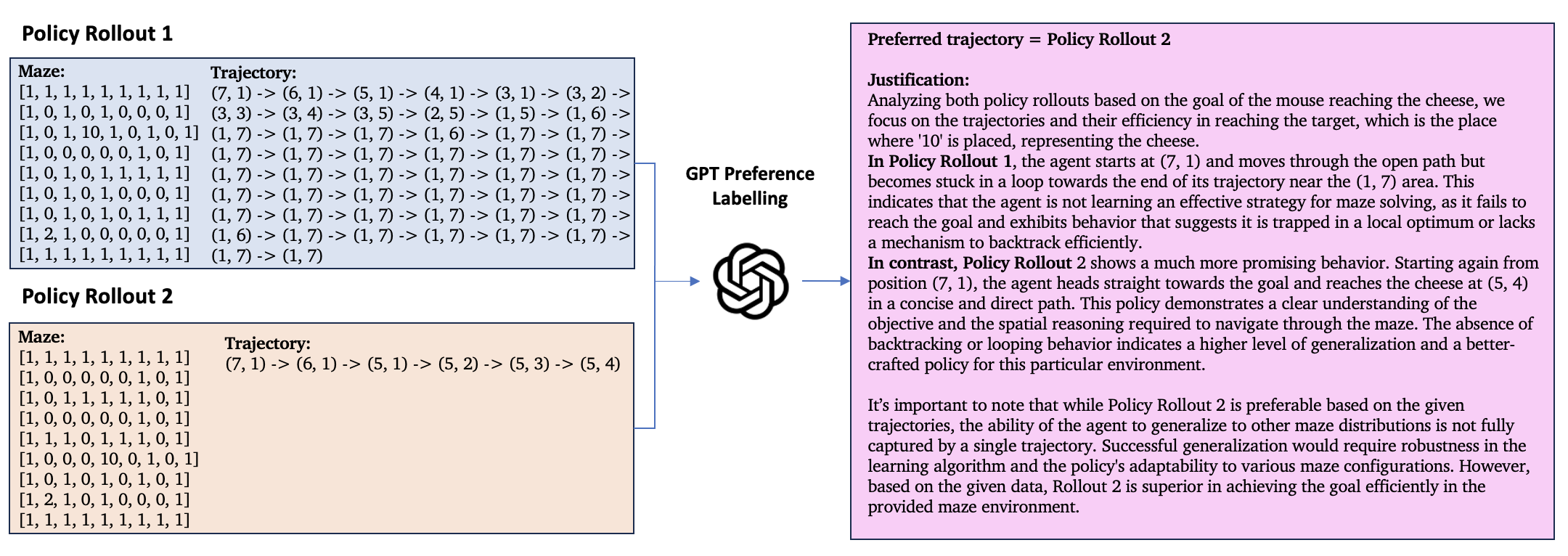}
  \caption{An example of a GPT-4 preference labeling.}
  \label{fig:llm_preference}
\end{figure}

We prompt GPT-4 with two candidate policy roll-outs and  instruct it to rate which policy roll-out is preferred. An example of the input to GPT-4 for this preference labeling is provided in figure \ref{fig:prompt_preference}).
An example of GPT-4 preference labeling is provided in Figure \ref{fig:llm_preference}. \\ 

\textit{Position Bias.} \hspace{0.1cm} \cite{lee_2023} and \cite{pezeshkpour} show that the order in which candidates are shown to LLM can influence its choice. This position bias is evident in our dataset as well: GPT-4 prefers choices in the second position in 56\% of the pairs, while it prefers choices in the first position in 43\% percent of pairs. \footnote{GPT is indifferent between both options 1\% percent of the times.} To address this position bias, we elicit two preferences for every pair of roll-out candidates. We switch the order in which roll-out candidates are presented to GPT-4 in each instance. We then average the results from both inferences to deduce the final preference distribution. 

\begin{figure}[h!]
  \centering
  \includegraphics[width = 1 \textwidth]{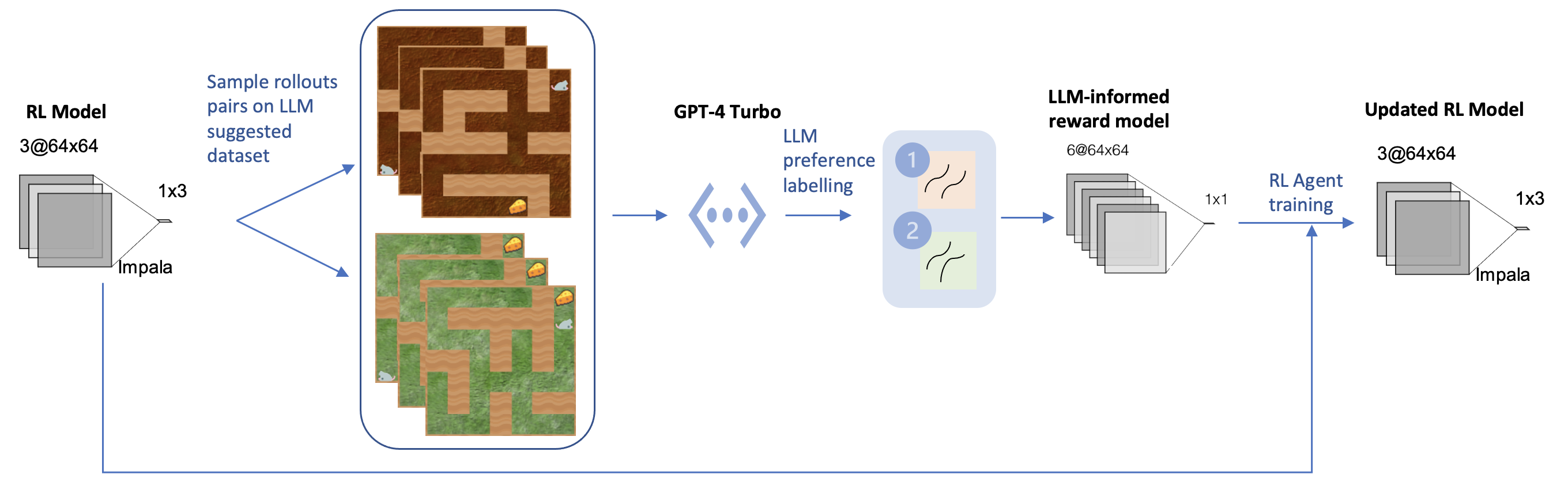}
  \caption{LLM preference modelling and reward model. The RL agent is deployed on the LLM generated dataset and its rollouts are stored. The LLM compares pairs of rollouts and provides preferences, which are used to train a new reward model. The reward model is then integrated to the remaining training timesteps of the agent.}
  \label{fig:methodology_step2}
\end{figure}

\paragraph{Reinforcement Learning with LLM feedback} From this preference distribution, we train a reward model as detailed in section \ref{sec:method_llm_reward_model}. The implementation details can be found in Appendix \ref{sec:appdx_reward_model}. 
We integrate the GPT-induced reward model and resume the training of the RL agent. We then implement the reinforcement learning using the Proximal Policy Optimization (PPO) algorithm (see Appendix \ref{sec:rl-ppo} for more details). A summary of this second step is provided in  Figure \ref{fig:methodology_step2}).

\subsection{Evaluation}
\paragraph{Reward Model from LLM Feedback}
To evaluate the reward model performance, we compute the Precision, Recall and F1 scores of our trained model when compared with the LLM labels. 
\paragraph{Performance of the RL agent} 
To assess the RL agent's performance, we use two metrics. The first, \textit{Capability}, is given by the training reward and serves as an indicator of the effectiveness of a reinforcement learning methodology in training the agent to achieve the intended task. The second, \textit{Generalization Capacity} is given by the test reward in unseen out-of-distribution scenarios and serves as an indicator of the agent's adaptability to unseen settings.

\section{Experimental results}

\subsection{Reward model}
We evaluate the Precision, Recall and F1 score for each class on the validation dataset. For each rollout pairs, the LLM preferences are in: $\{ (0, 1), (0.25, 0.75), (0.5, 0.5), (0.75, 0.25), (1, 0)\}$ and represents the strength in preference of rollout 1 or rollout 2. However, due to the very low prevalence of $(0.25, 0.75)$ and $(0.75, 0.25)$ preferences, each constituting only $0.8\%$, we present our metrics both as an average over all labels and specifically for the predominant classes: ${ (0, 1), (0.5, 0.5), (1, 0)}$. in Table \ref{tab:reward_model_metrics}.

\begin{table}[ht]
\caption{Classification metrics for the LLM-informed reward model on the validation rollout pairs}
\centering
\begin{tabular}{cccc}
\toprule
Class & Precision & Recall & F1 \\ \midrule
All 5 classes macro-average & 0.46 & 0.37 & 0.38 \\
(1.0, 0.0) & 0.95 & 0.52 & 0.67 \\
(0.5, 0.5) & 0.39 & 0.80 & 0.52 \\
(0.0, 1.0) & 0.97 & 0.53 & 0.69 \\
\bottomrule
\end{tabular}
\label{tab:reward_model_metrics}
\end{table}
Table \ref{tab:reward_model_metrics} shows that macro-average metrics are significantly influenced by the low-prevalence classes \footnote{See \ref{sec:appdx_reward_model} for more details on the classes.}. The high precision but lower recall for the $(0, 1)$ and $(1, 0)$ classes indicates that the LLM-reward model is often cautious in preferring one rollout over another, but accurate when it does. This tendency is beneficial, as it helps filter noise from the LLM labeling. Indeed, when the LLM-reward model drives the RL agent's training, even though the occurrences of high reward are rare, they are informative.

\subsection{RL Agent's Performance: Capacity and Goal Generalization}
\begin{figure}[h!]
  \centering
  \includegraphics[width = 1 \textwidth]{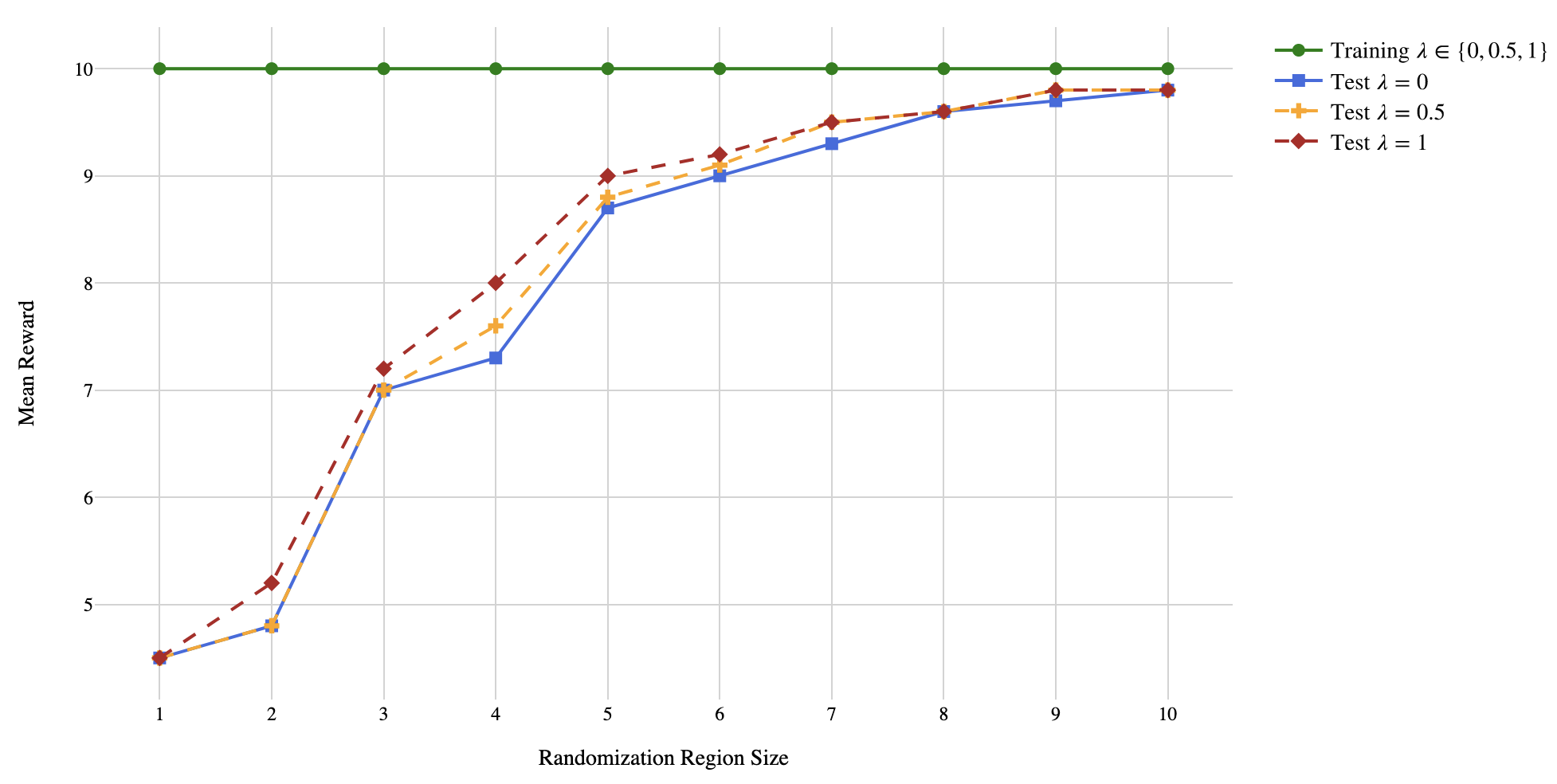}
  \caption{To provide a unique mean reward value over the training and test sets, we average the reward over the final $1.5M$ timesteps of the agent training. .}
  \label{fig:results_validation_rew}
\end{figure}

We assess our method on goal randomization regions ranging from size $0$ to size $10$.
Our method effectively reduces goal misgeneralization across 7 regions and matches the performance of the existing reward function in three others, as shown in \ref{fig:results_validation_rew}.

Our methodology consistently produces an agent that acts  \textit{capably} accross all randomization regions, as it always achieves the maximum mean training reward (10). 
For randomization regions $\{2,3,4,5,6,7,9\}$ our method effectively increases the agent's \textit{generalization capacity} as can be seen from its test reward being greater than with the environment reward function. 
The success of our method relies on the fact that the LLM-induced reward model captures a penalty for when the agent consistently navigates to the upper right corner despite the cheese being located elsewhere. Our method incentivizes exploration and better environmental cue recognition thus leading to better generalization in out-of-distribution settings. \\ 

The most pronounced improvement is observed in randomization region $4$, where the agent's learning is still flexible, allowing the reward model to more effectively influence its behavior. In contrast, with a random region size of 1, where intended and proxy goals are indistinguishable, the LLM-informed reward model ($\lambda=1$) does not enhance generalization. The disparity in test performance also lessens with larger random regions as the base agent independently learns a generalizable goal. \\ 

For training scenarios with $\lambda=0.5$ in randomization regions smaller than 4, our method matches the environmental reward function in terms of generalization. In these cases, the intended and proxy goals are hard to distinguish so the LLM-informed model's infrequent and mild penalties for biased behaviors are overshadowed by the environmental reward. However, this disparity fades with larger random regions, where the LLM-informed model becomes the primary driver of the agent's behavior. 

\section{Related Work}
\paragraph{Instances of Goal Misgeneralization and potential mitigations}  \cite{shah_2022} provide a broader definition of goal misgeneralization that apply to arbitrary learning systems and provide various examples across deep learning. \cite{langosco_2023}, provide 3 examples of goal misgeneralization for specific RL agents: the CoinRun Example, The Keys and Chest example, and the Maze example. Their mitigation solution consists in increasing the training dataset, yielding a more general agent policy. We use their Maze example as the basis to evaluate our method for reducing goal misgnerealization. \cite{song_2018} train their model on unrestricted adversarial examples to elicit and penalize misaligned behavior, while \cite{ziegler_2022} and \cite{perez_2022} generate examples to reduce the probability of unwanted language output. 

\paragraph{Scalable oversight} As models become more capable and their outputs become more complex, it will be increasingly harder to have humans supervise them (\cite{amodei_2016}). Consequently, it is necessary to think of robust techniques where the need for human input is minimized. Empirical work on scalable oversight is complicated as we currently do not have systems that can exceed our capabilities (\cite{bowman_2022}). Proposed scalable oversight techniques have included debate \cite{irving_2018}, self-critique (\cite{saunders_2022}, \cite{leike_2018})  or market making (\cite{hubinger_2020}).  
 
\paragraph{RLHF \& RLAIF.} Our method can be thought of as an extension of RLHF (\cite{christiano_2017}, \cite{stiennon_2022}, \cite{bai_2022}). Instead of relying on humans to convey the intended goal to the agent, we instead rely on a large language model to provide diversified training data and to reward the agent during the training. This is part of A growing field where other AI models are used to train the primary AI model. \cite{zhao_2021} and \cite{saunders_2022}’s work involve a model self-critiquing itself with natural language feedback and \cite{shi_2022}, involve work where the model is self-supervised. \cite{bai_2022_constitutional} implement a model of RLAIF (RL from AI feedback) where an AI model is trained on capturing human preferences from a construction and is then used for RL fine-tuning. In a similar fashion, \cite{klissarov2023motif} use LLM-preferences to train an RL agent on the  procedurally-generated NetHack game. They also explore combining the LLM based reward and the environment reward, achieveing greater task performance. Finally \cite{lee_2023} conduct a comparison of RLHF with RLAIF and find that they result in similar improvements over a baseline supervised fine tuned model.

\section{Discussion}
\subsection{LLM supervision capabilities}
One of the key foundations of our method relies on the fact that the LLM parent is able to assess the behavior of the main agent on two dimensions: its ability to fulfill the intended goal, and its capacity to generalize in different environments. Hence the LLM needs to be able to assess whether a particular policy acts toward the intended goal as well as hypothesize on potential aspects it could fail under distributional shifts. In particular, GPT-4 needs to be able to verify whether a particular trajectory could get the agent to the goal in the specific maze environment. In 77\% of the times, GPT-4 was correctly able to assess whether a specific trajectory was getting to the goal. \footnote{Value obtained by providing 100 examples of mazes and trajectories and asking GPT-4 to verify whether the agent reached the goal. 50 of these examples were trajectories that reached the goal.} This contrasts with the 10\% of times GPT-4 was able to solve the maze itself. \footnote{Value obtained by providing 50 examples of mazes to GPT-4, and prompting it to solve the maze. All the successes were on mazes that required 3 or less moves.} This highlights that, in our case, for LLM supervision, it suffices that the parent LLM be able to assess a specific policy relative to an intended goal, without necessarily being capable of the action itself. This aspect is particularly promising for scalable oversight: LLM can supervise other AI agents towards achieving their tasks and can help them generalize their goals even if they are not capable to do the tasks themselves.

\subsection{Environment Confoundedness vs Reward Model Informativeness}
As the randomization region grows, the agent is less biased towards navigating to the upper-right corner, and hence our method’s impact on correcting that bias diminishes. On the other hand, for small randomization regions, the goal is indistinguishable from its position, and hence the behavior of the agent cannot be corrected. Without a change in the training dataset or in the model architecture, the agent will confound the goal. It is interesting to note that regardless of the training method, the agent always learns a location rather a feature. This position-bias could happen for many reasons. First, it could be that learning based on a position may be simpler and more efficient than learning based on some feature. Second, it could be that the goal is not salient to the agent, and the positional information is more prominently represented and easier to interpret. Finally, given that the state space inherently provides more distinct cues about position in the form of coordinates, the learning could naturally prioritize position.  

\section{Conclusion}
In this paper, we have presented an approach to mitigate goal misgeneralization in reinforcement learning by integrating feedback from LLMs. Our method addresses the critical issue of inner alignment failure, where an RL agent confuses a proxy goal for the true goal during training, leading to suboptimal performance in out-of-distribution scenarios. By employing LLMs to analyze and refine RL agents' we demonstrated significant improvements in goal generalization, particularly in complex environments where proxy and true goals are distinct and behavioral biases are evident. This research suggests the potential of language models in providing scalable AI oversight and promoting goal-focused learning. Our findings open new avenues for future research in enhancing the robustness and adaptability of RL systems, paving the way for more reliable and efficient AI applications.

\newpage

\bibliographystyle{plainnat} 
\bibliography{bibliography}

\newpage 

\appendix

\section{Prompt \& GPT Analyses Examples} 
\begin{figure}[h!]
  \centering
  \includegraphics[width = 1 \textwidth]{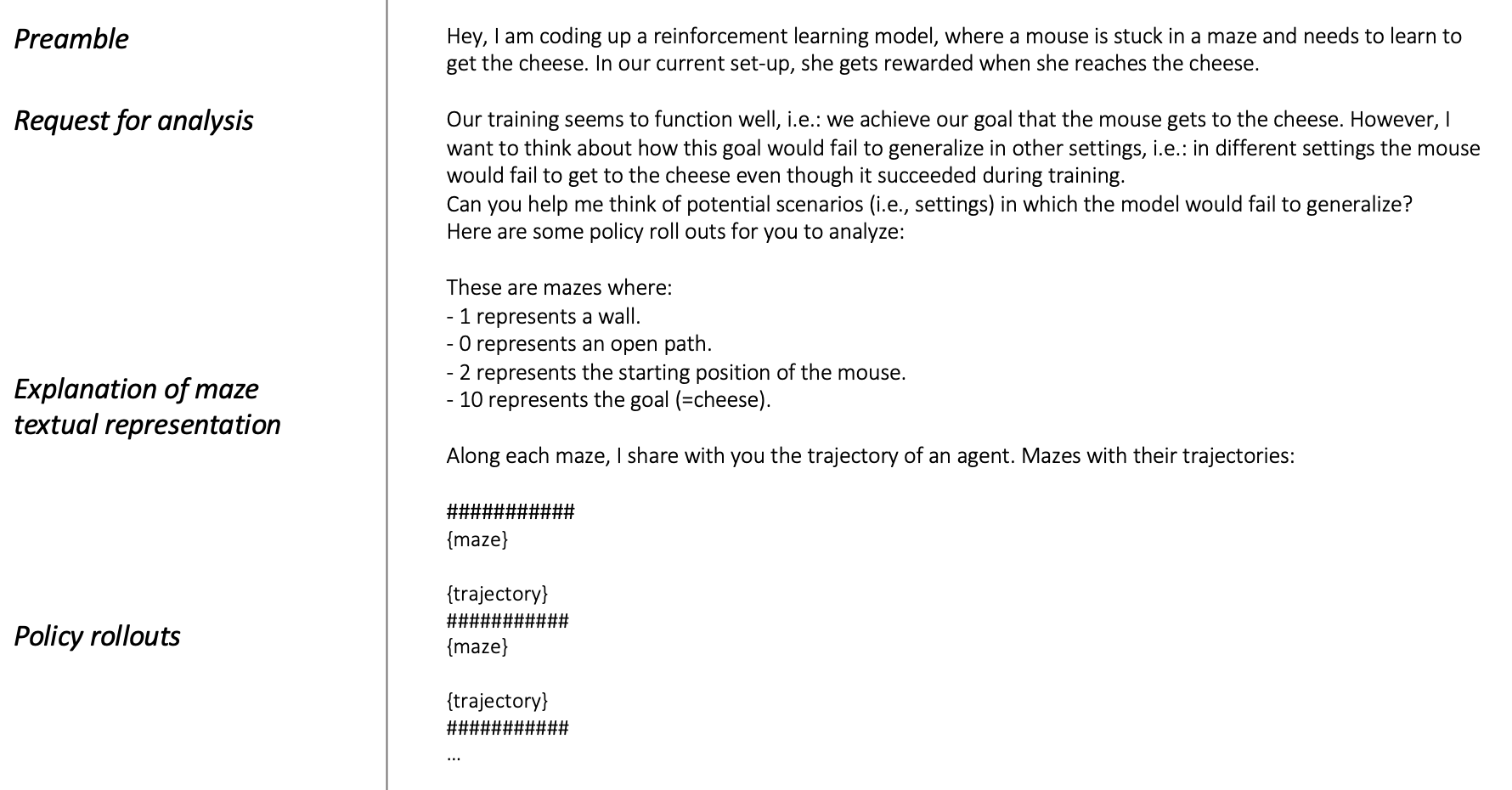}
  \caption{An example of a prompt provided to GPT-4 to examine the potential failures of the initially learnt policy. \{maze\} and \{trajectory\} are given examples sampled from the training dataset distribution.}
  \label{fig:prompt_env_analysis}
\end{figure}

\begin{figure}[h!]
  \centering
  \includegraphics[width = 1 \textwidth]{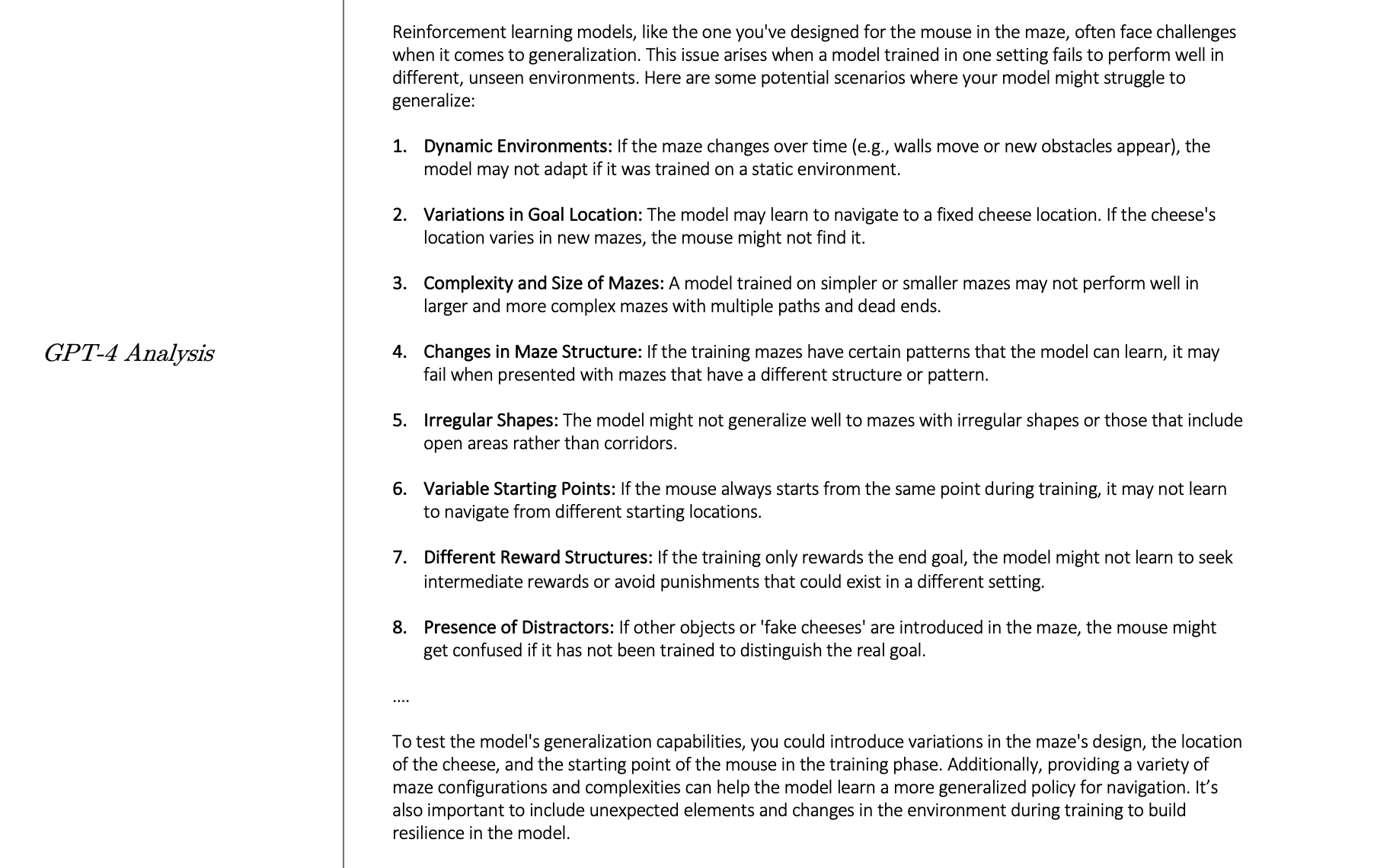}
  \caption{An example of GPT's answers when prompted to analyze the potential  failures of the initially learnt policy, see an example in Figure \ref{fig:prompt_env_analysis}}. 
  \label{fig:gpt_env_analysis}
\end{figure}

\begin{figure}[h!]
  \centering
  \includegraphics[width = 1 \textwidth]{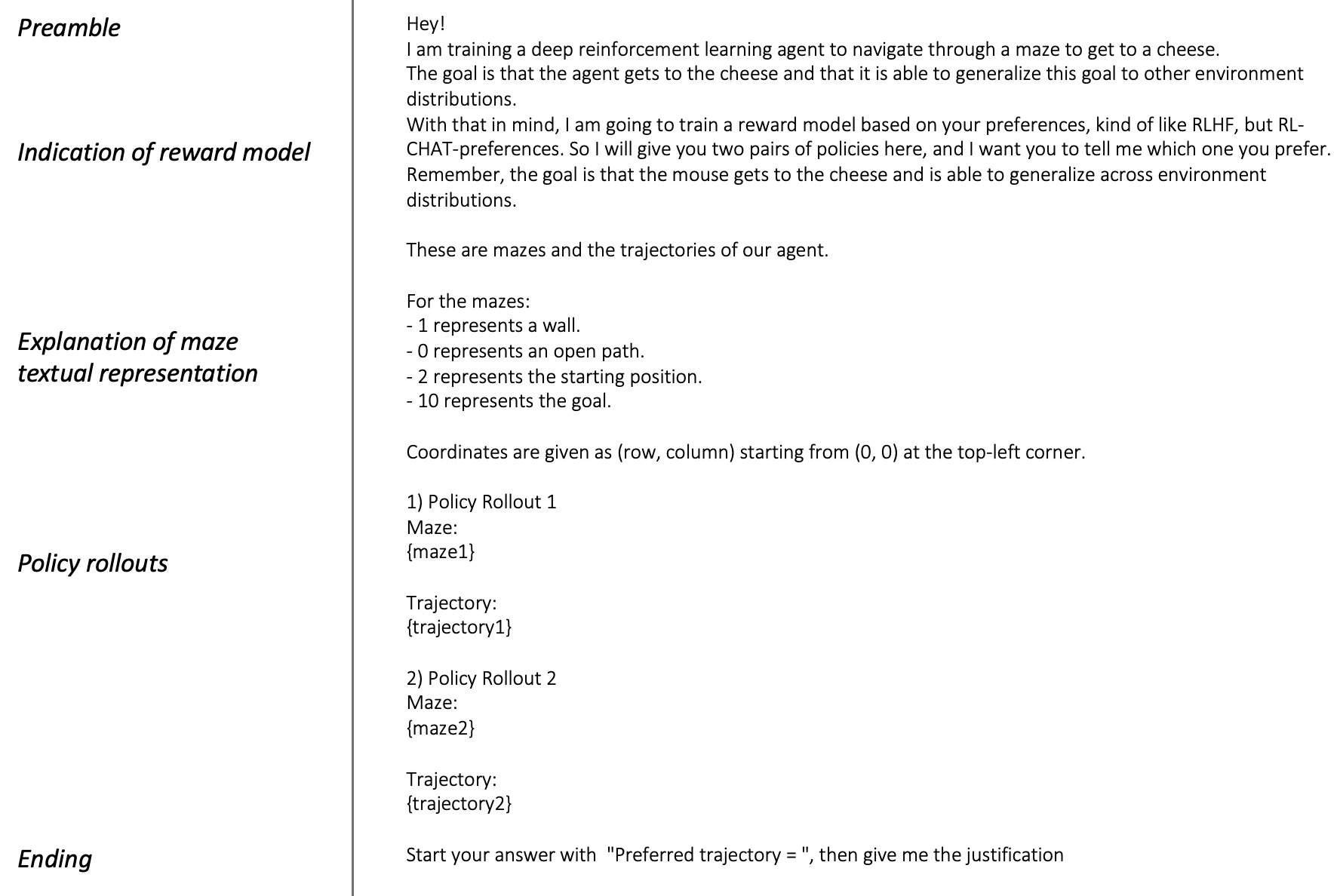}
  \caption{An example of a prompt provided to GPT-4 to give its preference over two policy rollouts. \{maze1\}, \{maze2\}, \{trajectory1\} and \{trajectory2\} are given examples sampled from the dataset suggested by the LLM.}
  \label{fig:prompt_preference}
\end{figure}

\section{Model Training Details}
\subsection{Reward model details} \label{sec:appdx_reward_model}
In $46\%$ of these roll-outs, the agent was able to get to the cheese within $50$ steps. Thus, to simplify the preference labeling and reward modeling, we cut all trajectories to have at most $50$ steps. 
We split the 158 mazes into 142 training and 16 validation. The training is performed on 1412 rollout pairs comparing rollouts uniquely from the 142 training mazes. The validation is performed on the remaining 388 pairs, each of which includes at least on rollout from the validation mazes. 
We train the reward model for 300 epochs with a batch size of 256, in 1 hour on an NVIDIA A-6000 GPU. At that point, the binary cross-entropy loss has plateaued. The CNN architecture of the reward model is the table \ref{tab:llm_rm} and was inspired by \cite{christiano_2017}. We use the Adam optimizer \citep{kingma2017adam} with a learning rate of $10^{-5}$.

\begin{table}[ht]
\centering
\caption{Hyper-parameters of the LLM-informed reward model.}
\begin{tabular}{lccccl}
\toprule
Layer & Output Channels/Features & Kernel Size & Padding & Stride  & Activation \\ \midrule
Conv2D & 16 & (5, 5) & 0 & 2 & \multicolumn{1}{c}{Leaky ReLU} \\
Conv2D & 16 & (3, 3) & 1 & 1 & \multicolumn{1}{c}{Leaky ReLU} \\ 
Conv2D & 16 & (5, 5) & 0 & 2 & \multicolumn{1}{c}{Leaky ReLU} \\ 
Conv2D & 16 & (3, 3) & 1 & 1 & \multicolumn{1}{c}{Leaky ReLU} \\ 
Conv2D & 16 & (3, 3) & 0 & 1 & \multicolumn{1}{c}{Leaky ReLU} \\ 
Conv2D & 16 & (3, 3) & 1 & 1 & \multicolumn{1}{c}{Leaky ReLU} \\ 
Conv2D & 16 & (3, 3) & 0 & 1 & \multicolumn{1}{c}{Leaky ReLU} \\ 
Conv2D & 16 & (3, 3) & 1 & 1 & \multicolumn{1}{c}{Leaky ReLU} \\ 
Linear & 64 & - & - & - & \multicolumn{1}{c}{Leaky ReLU} \\ 
Linear & 1 & - & - & - & \multicolumn{1}{c}{-} \\ 
\bottomrule
\end{tabular}
\label{tab:llm_rm}
\end{table}

Our model predicts probabilities of preference for rollouts 1 and 2. To transform these probabilities into classes, we use the following intervals presented in Table \ref{tab:pb_to_class}.

\begin{table}[h]
\centering
\caption{Preference probabilities to class.}
\begin{tabular}{cccccc}
\toprule
Class & (1.0, 0.0) & (0.75, 0.25) & (0.5, 0.5) & (0.25, 0.75) & (0.0, 1.0) \\ \midrule
Probability Rollout 1  & ].875, 1.0] & ].625, .875] & ].375, .625] & ].125, .375] & [0.0, .125] \\
Probability Rollout 2 & [0.0, .125] & ].125, .375] & ].375, .625] & ].625, .875] & ].875, 1.0] \\
\bottomrule
\end{tabular}
\label{tab:pb_to_class}
\end{table}

The Precision, Recall and F1 scores are shown in Table \ref{tab:reward_model_metrics_other_classes} for completion.

\begin{table}[h]
\centering
\caption{Hyper-parameters of the LLM-informed reward model.}
\begin{tabular}{cccc}
\toprule
Class & Precision & Recall & F1 \\ \midrule
(0.75, 0.25) & 0.00 & 0.00 & 0.00 \\
(0.25, 0.75) & 0.00 & 0.00 & 0.00 \\
\bottomrule
\end{tabular}
\label{tab:reward_model_metrics_other_classes}
\end{table}

\subsection{Reinforcement Learning with PPO}
\label{sec:rl-ppo}
Our experiments all use PPO, an actor-critic method. PPO is a reinforcement learning algorithm aimed to improve policy gradients methods by addressing their inefficiency and instability (\cite{schulman2017proximal}). 
In our implementation, the policy $\pi_{\theta}(a|s)$ is implemented by feedforward neural networks on top of a shared residual convolutional network with parameters $\theta$. All models are implemented with PyTorch and based on a codebase provided by \cite{Lee_2020}. 
We run a policy $\pi_{\theta_{\mathrm{old}}}$ in the environment for T timesteps. 
For each state-action pair, the advantage $\hat{A}_t$ is estimated, using Generalized Advantage Estimation (GAE). The objective function is defined as 
$$L^{CLIP}(\theta) = \hat{\mathbb{E}}_t (\min (r_t(\theta)\hat{A}_t, \textit{CLIP}(r_t(\theta), 1-\epsilon, 1+\epsilon)\hat{A}_t))$$
where $r_t(\theta)=\frac{\pi_{\theta}(a_t|s_t)}{\pi_{\theta_{\mathrm{old}}}(a_t|s_t)}$ is the probability ratio, and $\epsilon$ is a hyperparameter denoting the clipping range. \\
Policy updates are performed using Stochastic Gradient Ascent on the PPO objetive. The policy parameters $\theta$ are updated iteratively to maximize $L^{CLIP}(\theta)$. The objective function is maximized with K epochs and a minibatch size M. 
The hyperparameters are then the clip range $\epsilon$, the learning rates for policy and value networks, the number of epochs per update K, and the mini-batch size M.

\subsection{Reinforcement Learning Model Details}

We choose the same hyperparameters as \cite{langosco_2023}, from which we also base Table \ref{tab:hyperparameters}, to allow for direct comparison. The model is based on the Impala architecture \citep{espeholt2018impala}, without the recurrent components.

\begin{table}[ht!]
\centering
\caption{Hyperparameters for the reinforcement learning}
\label{tab:hyperparameters}
\begin{tabular}{@{}lc@{}}
\toprule
\textbf{Hyperparameter}            & \textbf{Value}   \\ \midrule
ENV. DISTRIBUTION MODE             & Easy            \\
$\gamma$                           & 0.999             \\
$\lambda$                          & 0.95              \\
LEARNING RATE                      & 0.0005           \\
\# TIMESTEPS PER ROLLOUT           & 256              \\
EPOCHS PER ROLLOUT                 & 3                \\
\# MINIBATCHES PER EPOCH           & 8                \\
MINIBATCH SIZE                     & 2048             \\
ENTROPY BONUS ($k_h$)              & 0.01             \\
PPO CLIP RANGE                     & 0.2               \\
REWARD NORMALIZATION?              & Yes             \\
\# WORKERS                         & 4                \\
\# ENVIRONMENTS PER WORKER         & 32             \\
TOTAL TIMESTEPS                    & 50M           \\
ARCHITECTURE                       & Impala           \\
LSTM?                              & No               \\
FRAME STACK?                       & No              \\
\# ENVIRONMENTS                     & 100,000        \\
\bottomrule
\end{tabular}
\end{table}

\subsection{Base Agent Training} \label{appdx:agent_training_curve}

\begin{figure}[h!]
  \centering
  \includegraphics[width = 0.92 \textwidth]{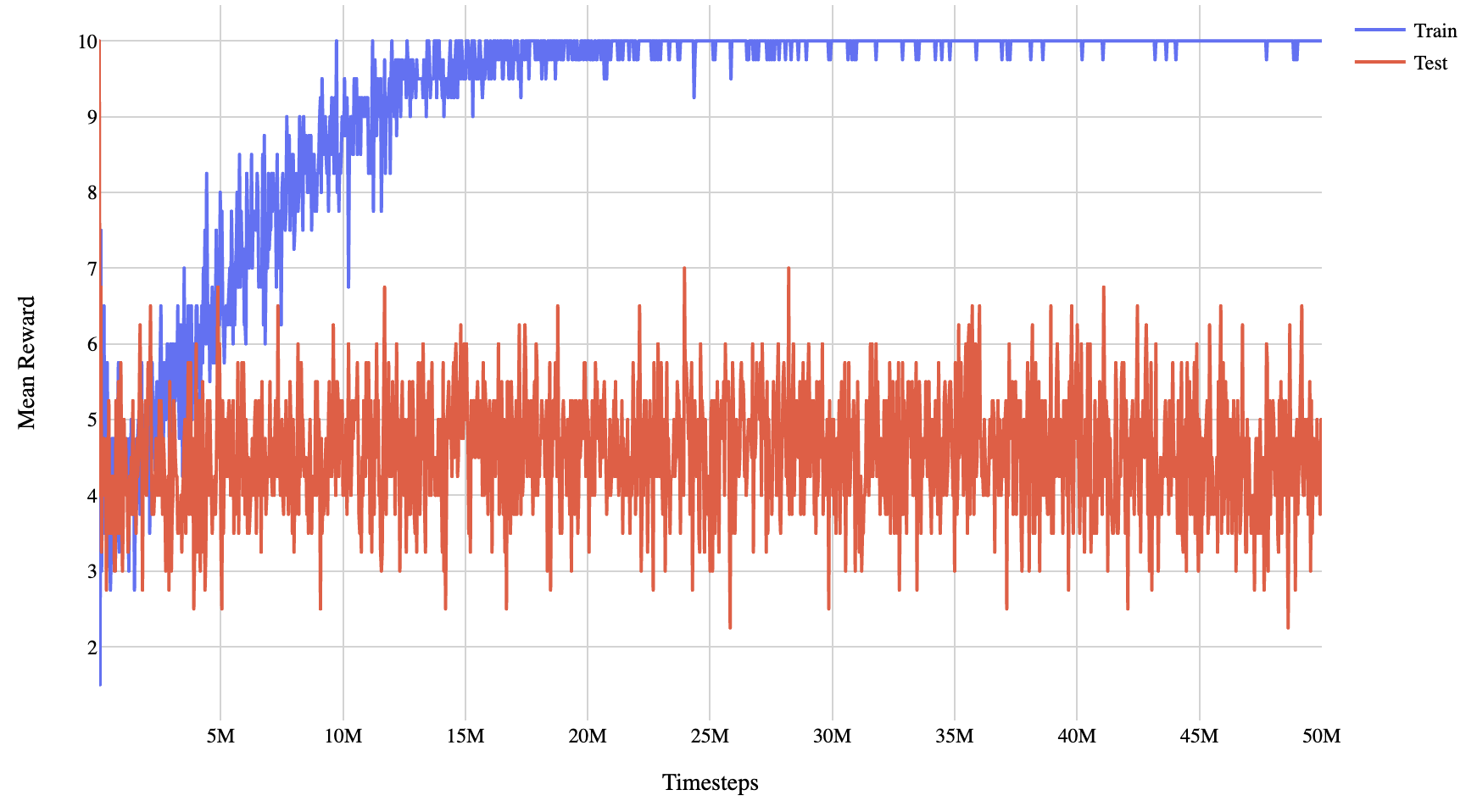}
  \caption{Mean reward on the train and test mazes for the base agent trained with the environment reward only, for $50M$ timesteps.}
  \label{fig:base_agent_training}
\end{figure}


\end{document}